\PassOptionsToPackage{unicode}{hyperref}
\PassOptionsToPackage{hyphens}{url}
\PassOptionsToPackage{dvipsnames,svgnames,x11names}{xcolor}
\documentclass[
]{article}
\usepackage{amsmath,amssymb}
\usepackage{lmodern}
\usepackage{iftex}
\ifPDFTeX
  \usepackage[T1]{fontenc}
  \usepackage[utf8]{inputenc}
  \usepackage{textcomp} 
\else 
  \usepackage{unicode-math}
  \defaultfontfeatures{Scale=MatchLowercase}
  \defaultfontfeatures[\rmfamily]{Ligatures=TeX,Scale=1}
\fi
\IfFileExists{upquote.sty}{\usepackage{upquote}}{}
\IfFileExists{microtype.sty}{
  \usepackage[]{microtype}
  \UseMicrotypeSet[protrusion]{basicmath} 
}{}
\makeatletter
\@ifundefined{KOMAClassName}{
  \IfFileExists{parskip.sty}{%
    \usepackage{parskip}
  }{
    \setlength{\parindent}{0pt}
    \setlength{\parskip}{6pt plus 2pt minus 1pt}}
}{
  \KOMAoptions{parskip=half}}
\makeatother
\usepackage{xcolor}
\usepackage{graphicx}
\makeatletter
\def\maxwidth{\ifdim\Gin@nat@width>\linewidth\linewidth\else\Gin@nat@width\fi}
\def\maxheight{\ifdim\Gin@nat@height>\textheight\textheight\else\Gin@nat@height\fi}
\makeatother
\setkeys{Gin}{width=\maxwidth,height=\maxheight,keepaspectratio}
\makeatletter
\def\fps@figure{htbp}
\makeatother
\NewDocumentCommand\citeproctext{}{}
\NewDocumentCommand\citeproc{mm}{%
  \begingroup\def\citeproctext{#2}\cite{#1}\endgroup}
\makeatletter
 \let\@cite@ofmt\@firstofone
 \def\@biblabel#1{}
 \def\@cite#1#2{{#1\if@tempswa , #2\fi}}
\makeatother
\newlength{\cslhangindent}
\newlength{\csllabelwidth}
\newenvironment{CSLReferences}[2] 
 {\begin{list}{}{%
  \setlength{\itemindent}{0pt}
  \setlength{\leftmargin}{0pt}
  \setlength{\parsep}{0pt}
  \ifodd #1
   \setlength{\leftmargin}{\cslhangindent}
   \setlength{\itemindent}{-1\cslhangindent}
  \fi
  \setlength{\itemsep}{#2\baselineskip}}}
 {\end{list}}
\usepackage{calc}

\ifLuaTeX
\usepackage[bidi=basic]{babel}
\else
\usepackage[bidi=default]{babel}
\fi
\babelprovide[main,import]{american}

\def\languageshorthands#1{}
\ifLuaTeX
  \usepackage{selnolig}  
\fi
\IfFileExists{bookmark.sty}{\usepackage{bookmark}}{\usepackage{hyperref}}
\IfFileExists{xurl.sty}{\usepackage{xurl}}{} 
\hypersetup{
  pdftitle={aweSOM: a CPU/GPU-accelerated Self-organizing Map and
Statistically Combined Ensemble Framework for Machine-learning
Clustering Analysis},
  pdfauthor={Trung Ha, Joonas Nättilä, Jordy Davelaar},
  pdflang={en-US},
  colorlinks=true,
  linkcolor={Maroon},
  filecolor={Maroon},
  citecolor={Blue},
  urlcolor={Blue},
  pdfcreator={LaTeX via pandoc}}

\title{\texttt{aweSOM}: a CPU/GPU-accelerated Self-organizing Map and
Statistically Combined Ensemble Framework for Machine-learning
Clustering Analysis}

\definecolor{c53baa1}{RGB}{83,186,161}
\definecolor{c202826}{RGB}{32,40,38}

\usepackage[affil-it]{authblk}
\usepackage{orcidlink}
\author[1,2,3%
  \ensuremath\mathparagraph]{Trung Ha%
    \,\orcidlink{0000-0001-6600-2517}\,%
    }
\author[4,5,2%
  ]{Joonas Nättilä%
    \,\orcidlink{0000-0002-3226-4575}\,%
    }
\author[6,7,2,5%
  ]{Jordy Davelaar%
    \,\orcidlink{0000-0002-2685-2434}\,%
    }

\affil[1]{Department of Astronomy, University of Massachusetts-Amherst,
Amherst, MA 01003, USA%
  }
\affil[2]{Center for Computational Astrophysics, Flatiron Institute, 162
Fifth Avenue, New York, NY 10010, USA%
  }
\affil[3]{Department of Physics, University of North Texas, Denton, TX
76203, USA%
  }
\affil[4]{Department of Physics, University of Helsinki, P.O. Box 64,
University of Helsinki, FI-00014, Finland%
  }
\affil[5]{Physics Department and Columbia Astrophysics Laboratory,
Columbia University, 538 West 120th Street, New York, NY 10027, USA%
  }
\affil[6]{Department of Astrophysical Sciences, Peyton Hall, Princeton
University, Princeton, NJ 08544, USA%
  }
\affil[7]{NASA Hubble Fellowship Program, Einstein Fellow%
  }
\affil[$\mathparagraph$]{Corresponding author: %
}
\date{08 October 2024}

\begin{document}
\maketitle

\section{Summary}\label{summary}

We introduce \texttt{aweSOM}, an open-source Python package for machine
learning (ML) clustering and classification, using a Self-organizing
Maps (SOM, \citeproc{ref-kohonen1990}{Kohonen, 1990}) algorithm that
incorporates CPU/GPU acceleration to accommodate large (\(N > 10^6\),
where \(N\) is the number of data points), multidimensional datasets.
\texttt{aweSOM} consists of two main modules, one that handles the
initialization and training of the SOM, and another that stacks the
results of multiple SOM realizations to obtain more statistically robust
clusters.

Existing Python-based SOM implementations (e.g., \texttt{POPSOM}, Yuan
(\citeproc{ref-yuan2018}{2018}); \texttt{MiniSom}, Vettigli
(\citeproc{ref-minisom}{2018}); \texttt{sklearn-som}) primarily serve as
proof-of-concept demonstrations, optimized for smaller datasets, but
lacking scalability for large, multidimensional data. \texttt{aweSOM}
provides a solution for this gap in capability, with good performance
scaling up to \(\sim 10^8\) individual points, and capable of utilizing
multiple features per point. We compare the code performance against the
legacy implementations it is based on, and find a \(10 - 100 \times\)
speed up, as well as significantly improved memory efficiency, due to
several built-in optimizations.

As a companion to this paper, Ha et al. (\citeproc{ref-ha2024}{2024})
demonstrates the capabilities of \texttt{aweSOM} in analyzing the
physics of plasma turbulence. Detailed instructions on how to install,
test, and replicate the results of the paper are available in the online
\href{https://awesom.readthedocs.io/en/latest/}{documentation}. Also
included in the documentation is an example of applying \texttt{aweSOM}
to the Iris dataset (\citeproc{ref-iris53}{Fisher, 1936}).

\section{Statement of need}\label{statement-of-need}

\subsection{The self-organizing map
algorithm}\label{the-self-organizing-map-algorithm}

A SOM algorithm is an unsupervised ML technique that excels at
dimensionality reduction, clustering, and classification tasks. It
consists of a 2-dimensional (2D) lattice of nodes. Each node contains a
weight vector that matches the dimensionality of the input data. A SOM
performs clustering by adapting the weight vectors of nodes,
progressively reshaping the lattice's topology to match the intrinsic
clustering of the input data. In this manner, a SOM lattice can capture
multidimensional correlations in the input data.

SOM is commonly used in various real-world applications, such as in the
financial sector (e.g., \citeproc{ref-Alshantti2021}{Alshantti \&
Rasheed, 2021}; \citeproc{ref-Pei2023}{Pei et al., 2023}), in
environmental surveys (e.g., \citeproc{ref-Alvarez2008}{Alvarez-Guerra
et al., 2008}; \citeproc{ref-Li2020}{Li et al., 2020}), in medical
technology (e.g., \citeproc{ref-Hautaniemi2003}{Hautaniemi et al.,
2003}; \citeproc{ref-Kawaguchi2024}{Kawaguchi et al., 2024}), among
others. \texttt{aweSOM} is originally developed to be used in analyzing
astrophysical simulations, but can be applied to a wide variety of
real-world data.

\subsubsection{\texorpdfstring{\texttt{POPSOM}}{POPSOM}}\label{popsom}

We base the SOM module of \texttt{aweSOM} on \texttt{POPSOM}
(\citeproc{ref-hamel2019}{Hamel, 2019}; \citeproc{ref-yuan2018}{Yuan,
2018}), a R-based SOM model. \texttt{POPSOM} was developed as a
single-threaded, stochastic training algorithm with built-in
visualization capabilities. However, due to its single-threaded nature,
the algorithm does not scale well with large datasets. When
\(N \gtrsim 10^6\), \texttt{POPSOM} is often unable to complete the
training process as the dimensionality of the input data increases due
to its much higher memory usage. As an example, we generated a mock
dataset with \(N = 10^6\) and \(F = 6\) dimensions, then trained it on a
lattice of \(X = 63\), and \(Y = 32\), where \(X, Y\) are the dimensions
of the lattice, using one Intel Icelake node with 64 cores and 1 TB
memory. \texttt{POPSOM} completed the training in \(\approx 2200\) s and
consumed \(\approx 600\) GB of system memory at its peak.

\subsubsection{\texorpdfstring{Rewriting \texttt{POPSOM} into
\texttt{aweSOM}}{Rewriting POPSOM into aweSOM}}\label{rewriting-popsom-into-awesom}

To combat the long training time and excessive memory usage, we rewrite
\texttt{POPSOM} with multiple optimizations/parallelizations. We
replaced legacy code with modern \texttt{NumPy} functions for updating
the lattice (a 3D array) and eliminated the use of \texttt{pandas}
DataFrames (\citeproc{ref-pandas}{The pandas development team, 2024}),
which consume significantly more memory. The weight vector modifications
in the DataFrame were also less efficient compared to the NumPy arrays
used in aweSOM. Furthermore, for the steps where parallelization could
be leveraged (such as when the cluster labels are mapped to the lattice,
then to the input data), we integrate \texttt{Numba}
(\citeproc{ref-numba}{Lam et al., 2015}) to take advantage of its
Just-In-Time (JIT) compiler and simple parallelization of loops. In the
same example as above, \texttt{aweSOM} took \(\approx 200\) s and
consumed \(\approx 450\) MB of memory to complete the training and
clustering. In addition to the \(\sim 10 \times\) speedup,
\texttt{aweSOM} is also \(\sim 10^3 \times\) more memory-efficient.

The left hand side of \autoref{fig:sce_scaling} shows a graph of the
performance between \texttt{aweSOM} and the legacy \texttt{POPSOM}
implementation over a range of \(N\) and \(F\), performed on one Intel
Icelake compute node with 64 CPU cores and 1 TB memory. While
\texttt{POPSOM} initially performs slightly faster than \texttt{aweSOM}
for \(N \lesssim 10^4\), this changes when \(N\) exceeds
\(5 \times 10^5\), after that \texttt{aweSOM} consistently outperforms
\texttt{POPSOM} by approximately a factor of \(10\). Critically,
\texttt{POPSOM} fails to complete its clusters mapping for
\(N \gtrsim 10^6, F > 4\) because the memory buffer of the test node was
exceeded.

\subsection{The statistically combined ensemble
method}\label{the-statistically-combined-ensemble-method}

The statistically combined ensemble (SCE) method was developed by Bussov
\& Nättilä (\citeproc{ref-bussov2021}{2021}) to stack the result of
multiple independent clustering realizations into a statically
significant set of clusters. This method represents a form of ensemble
learning. Additionally, SCE can also be used independently from the base
SOM algorithm, and is compatible with any general unsupervised
classification algorithm.

\subsubsection{The legacy SCE
implementation}\label{the-legacy-sce-implementation}

In its original version, the SCE was saved as a nested dictionary of
boolean arrays, each of which contains the spatial similarity index
\(g\) between cluster \(C\) and cluster \(C'\). The total number of
operations scales as \(N_{C}^R\), where \(N_C\) is the number of
clusters in each realization, and \(R\) is the number of realizations.
For example, in our use case involving plasma simulation data
(\citeproc{ref-ha2024}{Ha et al., 2024}), each SOM realization produces
on average 7 clusters, and the SCE analysis incorporates 36
realizations, resulting in approximately \(7^{36} \sim 10^{30}\)
array-to-array comparisons.

\subsubsection{\texorpdfstring{Integrating SCE into \texttt{aweSOM} with
\texttt{JAX}}{Integrating SCE into aweSOM with JAX}}\label{integrating-sce-into-awesom-with-jax}

To mitigate this bottleneck, we rewrite the legacy SCE code with
\texttt{JAX} (\citeproc{ref-jax}{Bradbury et al., 2018}) to
significantly enhance the performance of array-to-array comparisons
(which are matrix multiplications) by leveraging the GPU's
parallel-computing advantage over the CPU. We implement this
optimization by replacing the original nested dictionaries with data
arrays. Then, every instance of matrix operation using \texttt{NumPy} is
converted to \texttt{jax.numpy}. Additionally, we implement internal
checks such that the SCE code automatically reverts to \texttt{NumPy} if
GPU-accelerated \texttt{JAX} is not available.

\begin{figure}
\centering
\includegraphics{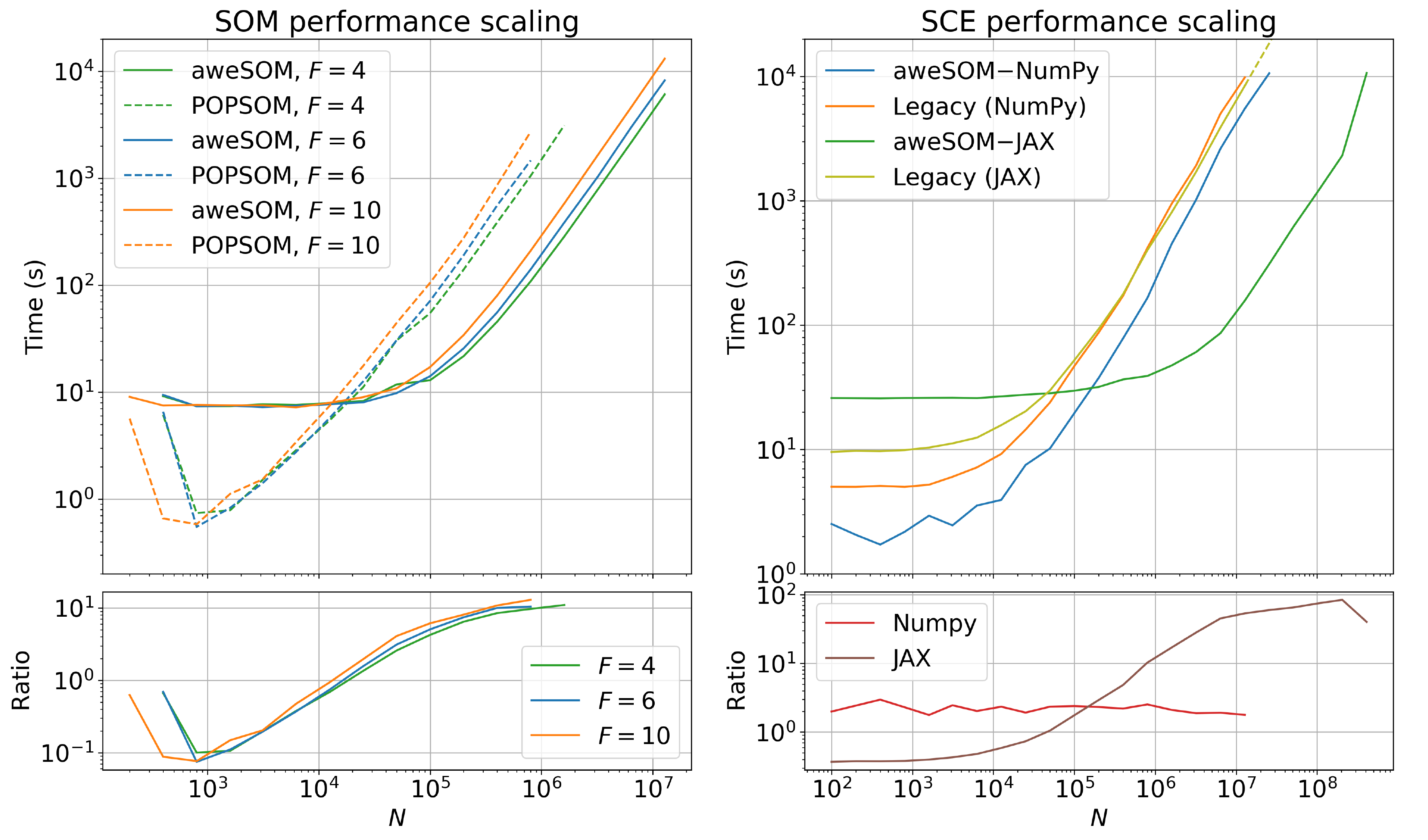}
\caption{Performance scaling for \texttt{aweSOM} vs.~the legacy SOM
(left) and SCE (right) implementation. The top panels show the time for
each implementation to complete analysis of \(N\) number of data points.
On the right panel, the dotted line extending from the olive line shows
linear extrapolations from the data in order to estimate the speedup.
The bottom panels show the ratio between the time taken by the legacy
code divided by the time taken by \texttt{aweSOM}. In the SOM analysis,
we consider a dataset with \(F = 6\) and \(F = 10\) dimensions. In the
SCE analysis, we test the scaling of both a GPU-accelerated
implementation (with \texttt{JAX}) and a CPU-only implementation (with
\texttt{NumPy}). \label{fig:sce_scaling}}
\end{figure}

Similar to the SOM implementation, the SCE implementation in
\texttt{aweSOM} demonstrates excellent scalability as the number of data
points increases. The right hand side of \autoref{fig:sce_scaling} shows
a graph of the performance between the two implementations given
\(R = 20\). At \(N < 5 \times 10^4\), the legacy code is faster due to
the overhead from loading \texttt{JAX} and the JIT compiler. However,
\texttt{aweSOM} quickly exceeds the performance of the legacy code, and
begins to approach its maximum speed-up of \(\sim 100 \times\) at
\(N \gtrsim 10^7\) (performed on one NVIDIA A100-40GB GPU). On the other
hand, when running on CPU-only with \texttt{NumPy}, \texttt{aweSOM}
consistently shows a \(2 \times\) speed improvement over the legacy
code. Altogether, it is best to use \texttt{aweSOM} with \texttt{Numpy}
when \(N \lesssim 10^5\), and with \texttt{JAX} when \(N \gtrsim 10^5\).

\section{Acknowledgements}\label{acknowledgements}

The authors would like to thank Kaze Wong for the valuable input in
setting up \texttt{JAX} for the SCE analysis. The authors would also
like to thank Shirley Ho and Lorenzo Sironi for the useful discussions.
TH acknowledges support from a pre-doctoral program at the Center for
Computational Astrophysics, which is part of the Flatiron Institute. JN
is supported by an ERC grant (ILLUMINATOR, 101114623). JD is supported
by NASA through the NASA Hubble Fellowship grant HST-HF2-51552.001-A,
awarded by the Space Telescope Science Institute, which is operated by
the Association of Universities for Research in Astronomy, Incorporated,
under NASA contract NAS5-26555. \texttt{aweSOM} was developed and
primarily run at facilities supported by the Scientific Computing Core
at the Flatiron Institute. Research at the Flatiron Institute is
supported by the Simons Foundation.

\section*{References}\label{references}
\addcontentsline{toc}{section}{References}

\phantomsection\label{refs}

\end{document}